%% file: paper.tex
\documentclass{article}

\usepackage{microtype}
\usepackage{graphicx}
\usepackage{subcaption}
\usepackage{booktabs} % for professional tables

\usepackage{hyperref}

\usepackage[preprint]{icml2026}

\usepackage[table]{xcolor}
\usepackage{pifont}
\usepackage{comment}
\usepackage{algorithm}
\usepackage{algorithmic}
\usepackage{amsmath}
\usepackage{amssymb}
\usepackage{mathtools}
\usepackage{amsthm}
\usepackage{multirow}
\usepackage{dcolumn}
\usepackage{booktabs}
\usepackage{graphicx}
\usepackage{subcaption}
\newcolumntype{d}{D{.}{.}{2}}
\DeclareMathOperator*{\argmin}{arg\,min}

\usepackage[capitalize,noabbrev]{cleveref}

\theoremstyle{plain}

\theoremstyle{definition}

\theoremstyle{remark}

\icmltitlerunning{MGVQ: Synergizing Multi-dimensional Sensitivity-Aware and Gradient-Hessian Fusion for Vector Quantization}

\makeatletter
\gdef\icmlcorrespondingauthor@text{Dawei Yang}
\makeatother

\begin{document}
\twocolumn[
  \icmltitle{MGVQ: Synergizing Multi-dimensional Sensitivity-Aware and Gradient-Hessian Fusion for Vector Quantization}

  \icmlsetsymbol{equal}{*}
  \icmlsetsymbol{corresponding}{\textdagger}

  \begin{icmlauthorlist}
    \icmlauthor{Zhong Wang}{equal,bmstu}
    \icmlauthor{Zukang Xu}{equal,houmo}
    \icmlauthor{Xing Hu}{houmo}
    \icmlauthor{Dawei Yang}{houmo,corresponding}
  \end{icmlauthorlist}

  \icmlaffiliation{houmo}{Houmo AI}
  \icmlaffiliation{bmstu}{Bauman Moscow State Technical University}
  \icmlkeywords{Large Vision-Language Models, Vector Quantization, Post-Training Quantization, Mixed-Precision Quantization}

  \vskip 0.3in
]

\printAffiliationsAndNotice{\icmlEqualContribution \textsuperscript{\textdagger}Corresponding author.}

\input{sec_0.paper}

\end{document}

%% file: sec_0.paper.tex
% 需要修改的点：
% 0.icml模板适配，消除编译错误。 √
% 1.introduction部分的讲述的梯度问题，应该改一下描述，不说是发现了梯度存在，而是说成做完误差补偿后，原始收敛点被改变，出现梯度。√
% 2.增加对YAQA的引用，说清楚是YAQA未考虑这种一阶项的影响。在introduction里写√
% 3.改一下文章题目和abstract的语言，和公开版本iclr避开。√
% 审稿人关键问题汇总：
% 4.图6画的更复杂一点，画的更加详细一点。 √
% 5.在4.1节中写的关键挑战是“如何分配有限位预算”，这块要改一下用词，不然和introduction里面提到的挑战冲突了。关键挑战改为注意点“如何分配有限位预算”√
% 6.关键步骤（∇L ≈ βE）（使用残差（E）作为代理梯度）缺乏对齐证据。没有对（cos(∇L, E)）的测量，没有对（|∇L - βE|）的边界限制，也没有针对（β）的逐层稳健性。在各向异性曲率的情况下，（E）可能指向低显著性方向，导致补偿错位。 加附录 证明参考GuidedQuant  √

\begin{abstract}
\input{sec_1.abstract}
\end{abstract}
\input{sec_2.introduction}
\input{sec_3.related_work}
\input{sec_4.preliminaries}
\input{sec_5.method}

\input{sec_6.experiments}

\vspace{-2mm}
\input{sec_7.conclusion}
\vspace{-2mm}

\clearpage
\input{sec_9.Impact_Statement}
\bibliographystyle{icml2026}
\bibliography{icml2026_conference}

\input{sec_8.appendix}

%% file: sec_1.abstract.tex
Vision-Language Models (VLMs) have demonstrated impressive capabilities. Nevertheless, their sheer scale presents a major obstacle to deployment on resource-constrained devices. Among compression strategies, vector quantization (VQ) emerges as a superior approach due to its high representational efficiency at ultra-low bitwidths. VQ operates by establishing a compact codebook where weight vectors are assigned to their nearest discrete codewords, thereby reducing memory footprint and bandwidth usage while maintaining the model's performance.
However, applying VQ directly to VLMs faces two fundamental challenges: 
(1) Modality-induced weight heterogeneity.
In VLMs, image and text inputs induce divergent weight distributions, which a unified codebook fails to capture.
(2) Error compensation mismatch from ignoring first-order gradients.
In VLMs, existing second-order error compensation methods shift weights from pre-trained convergence points, ignoring gradient drift and leading to inaccurate loss with biased compensation.
To overcome these hurdles, we introduce \textbf{MGVQ} (Synergizing \textbf{M}ulti-dimensional Sensitivity-Aware and \textbf{G}radient-Hessian Fusion for \textbf{V}ector \textbf{Q}uantization), a framework comprising two pivotal components:
(1) Sensitivity-driven structured
mixed-precision quantization, a mixed-precision scheme that allocates bit-widths based on channel sensitivity, combining global and local sensitivity metrics for fine-grained and interpretable resource distribution. 
(2) Gradient-aware second-order error compensation, a compensation method that explicitly incorporates first-order gradients to address their non-negligible role in VLM quantization errors, with efficient computation enabled by Kronecker and Block-LDL decompositions.
We validate MGVQ across leading VLMs, such as LLaVA-onevision, InternVL2, and Qwen2-VL. Under 2-bit quantization, our method consistently outperforms state-of-the-art PTQ baselines, delivering accuracy gains of up to \textbf{+4.9} (71.4\% vs. 67.0\% on InternVL2-26B). These results highlight MGVQ as a robust and efficient solution for ultra-low-bit quantization, facilitating the practical deployment of multimodal models in resource-constrained scenarios.

%% file: sec_2.introduction.tex
\section{Introduction}\label{sec_intro}

\begin{figure}[htbp]
  \centering
  \includegraphics[width=1\linewidth]{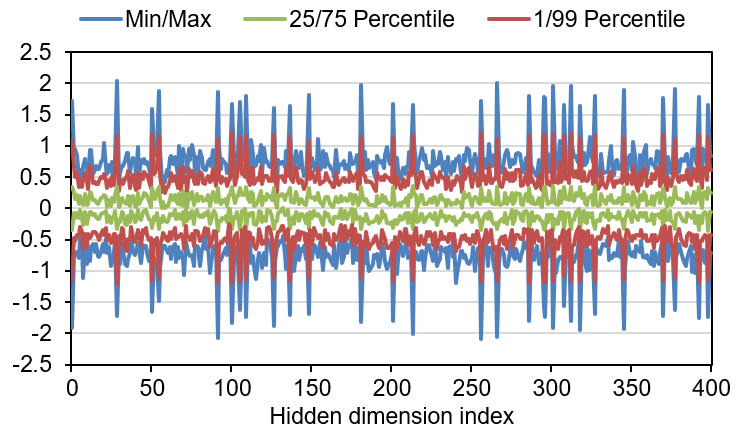}
  \caption{Weight distribution map of layer.1.down\_proj in LLaVA-OneVision-7B.}
  \label{img2_quantile_w}
\end{figure}

\begin{figure}[htbp]
  \centering
  \includegraphics[width=1\linewidth]{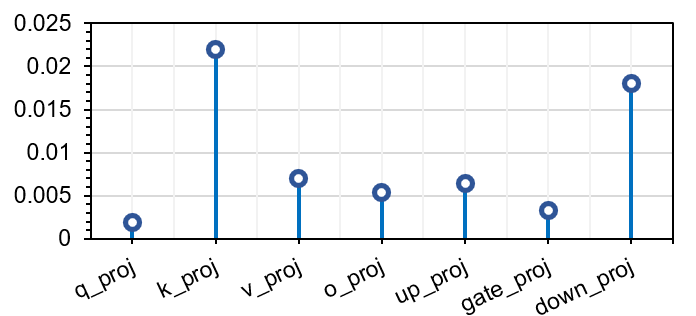}
  \caption{Gradient value at the 99\% quantile of the gradient statistics of the 31st block of Qwen2-VL-72B after YAQA.}
  \label{img3_m}
\end{figure}

Vision-Language Models (VLMs) are multimodal AI systems that integrate computer vision and natural language processing, taking both text and image/video inputs to generate text outputs, thereby enabling rich cross-modal reasoning and interaction \citep{AIVLM,zhang2024vision,liu2023visual,Qwen-VL,Qwen2-VL}. However, these models typically contain billions of parameters, making training and inference computationally expensive and limiting their deployment in latency-sensitive or resource-constrained environments \citep{VLMQ}. For instance, Qwen2-VL-72B requires over 140GB of GPU memory during the prefill stage under FP16 inference, far exceeding the capacity of most edge devices. Reducing memory and bandwidth requirements while maintaining accuracy is therefore essential for practical deployment of large VLMs.
%\begin{figure}[r]{0.5\textwidth}
\begin{figure}[t]
    \centering
    %\vspace{-8mm}
    \includegraphics[width=1\linewidth]{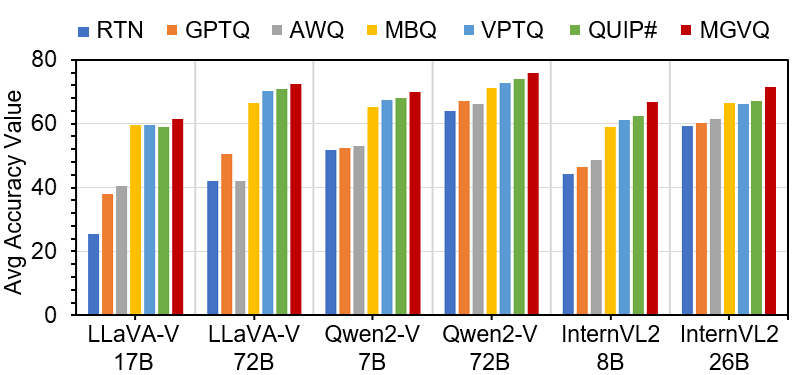}
    %\vspace{-6mm}
    \captionof{figure}{Comparison of average accuracy between MGVQ and other quantization methods across different VLMs}
    \label{img4_acc}
    \vspace{-4mm}
\end{figure}
%\vspace{-4mm}
% \textbf{Motivation for quantization.}
Post-training quantization (PTQ) avoids expensive retraining and substantially reduces storage and memory bandwidth, making it a key technique for compressing LLMs\citep {gptq, gptaq, MambaQuant}. Currently, PTQ methods can be broadly categorized into two classes. Scalar quantization (SQ), which performs well at medium to high bit-widths ($\geq 4$ bits), assigns each weight an independent scaling factor and zero point, offering a lightweight representation~\citep{gptq,awq,illm}.
Nevertheless, as the bitwidth decreases to 3 bits or lower, the representational capacity of SQ becomes severely limited, resulting in sharp accuracy degradation.
In contrast, vector quantization (VQ) maps high-dimensional weight vectors into a shared codebook, exploiting structural redundancy to achieve higher compression ratios \citep{vector_quantize}. This approach has been shown to substantially improve quantization performance under ultra-low-bitwidth settings \citep{gptvq, vptq, PCDVQ}.
\begin{figure}[t]
    % \vspace{-4mm}
    \centering 
    \includegraphics[width=1\linewidth]{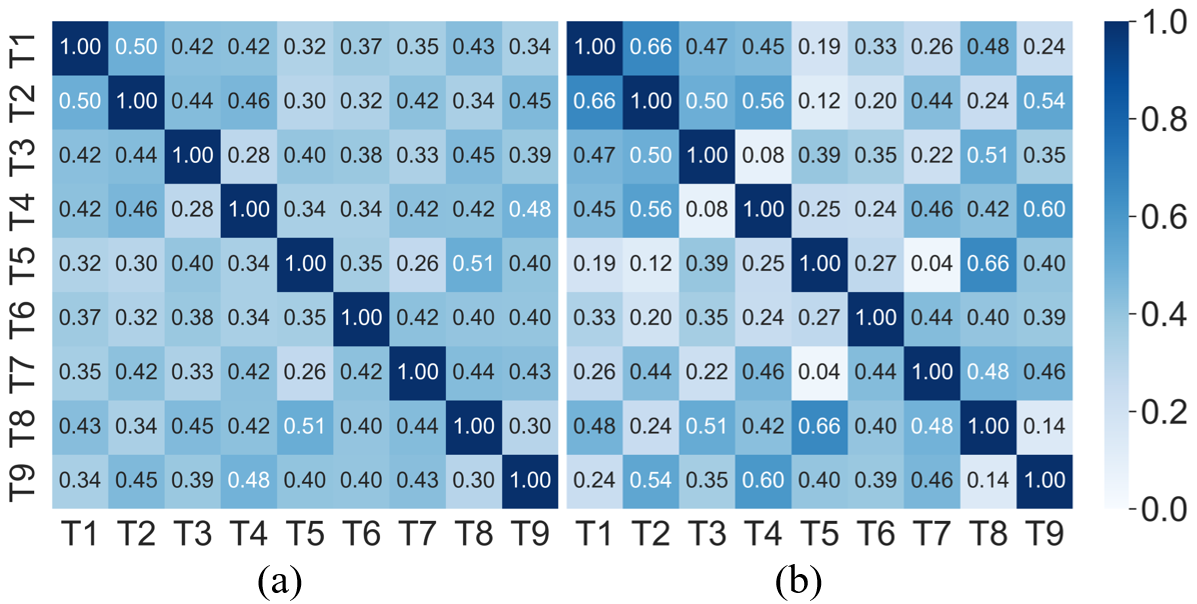}
    % \vspace{-2mm}
    \caption{Similarity between tokens. (a) Text tokens similarity. (b) Image tokens similarity}
    \label{img1_Similarity}
    % \vspace{-3mm}
\end{figure}
% \textbf{Challenges in VLM quantization.}
%\vspace{-6mm}
However, directly applying vector quantization to VLMs leads to severe accuracy degradation due to two fundamental challenges:
(1) Modality-induced weight heterogeneity. Within the same layer, VLM weights must simultaneously adapt to image and text tokens. As shown in Figure \ref{img1_Similarity}, these two types of tokens exhibit markedly different statistical characteristics, resulting in the heterogeneous weight distributions illustrated in Figure \ref{img2_quantile_w}. Applying a unified codebook or fixed bit allocation across an entire layer fails to accommodate such structural heterogeneity, thereby amplifying quantization errors.
(2) Error compensation mismatch from ignoring first-order gradients. Existing second-order error compensation methods inevitably shift weights away from their pre-trained convergence point. This displacement breaks the local optimality, inducing significant first-order gradients that were previously negligible, as illustrated in Figure \ref{img3_m}. By relying solely on Hessian curvature, conventional methods such as YAQA~\citep{YAQA} and VPTQ~\citep{vptq} neglect this gradient drift. Consequently, such an oversight causes inaccurate loss approximation and biased error compensation, ultimately degrading final model performance.
% \textbf{Our approach.}

To address these limitations, we propose MGVQ (Synergizing Multi-dimensional Sensitivity-Aware and Gradient-Hessian Fusion for Vector Quantization), a quantization framework consisting of two key components:
(1) Sensitivity-driven structured mixed-precision quantization (SSMQ). We integrate both global and local sensitivity metrics to partition sub-blocks of weights and allocate optimal bit under a fixed bit budget. Highly sensitive regions are assigned more bits, enabling fine-grained and interpretable resource allocation.
(2) Gradient-aware error compensation (GAEC). For each layer, we perform a Taylor expansion of the global loss, where the quantization residual is used to approximate the first-order gradient matrix and the second-order Hessian is approximated via Kronecker factorization. Based on this formulation, we derive the theoretically optimal compensation rule and apply it iteratively to progressively reduce quantization errors.

These two components enable channel-level adaptive bit allocation and error compensation under resource-constrained conditions, while overcoming the limitation of conventional PTQ methods that neglect first-order gradient terms. 
This significantly alleviates the accumulation of quantization errors in deep networks and their adverse impact during cross-modal propagation. 
Experimental results show that under 2-bit quantization, as observed in Figure \ref{img4_acc}, MGVQ consistently outperforms existing approaches across multiple representative VLMs. For example, on the InternVL\citep{internvl} model, MGVQ achieves more than a 4\% improvement over QuIP\# \citep{QuIP4PLUS}, and ablation studies further validate the independent contributions of each module.

% \textbf{Contributions}. 
The main contributions of this paper are summarized as follows:
\begin{itemize}
    \item We identify two VLM-specific challenges for vector quantization:  modality-induced weight heterogeneity and error compensation mismatch from ignoring first-order gradients.
    \item We propose MGVQ, which combines multi-dimensional sensitivity analysis, structured mixed-precision allocation, and gradient-aware error compensation to mitigate cross-layer and cross-modal quantization errors.
    \item We conduct extensive experiments on representative VLMs, showing that MGVQ achieves superior accuracy under low bitwidth quantization while maintaining efficiency, outperforming existing state-of-the-art (SOTA) methods.
\end{itemize}

%% file: sec_3.related_work.tex
\section{Related Work}\label{sec_related_work}
%\vspace{-3mm}
\textbf{Scalar Quantization (SQ)} maps parameters to uniformly spaced levels using a shared scaling factor and zero-point, implicitly assuming isotropy in the parameter space and uniform channel sensitivity. Current SQ schemes, when combined with auxiliary optimization techniques, have demonstrated strong performance at 4-bit precision and above. Representative approaches include GPTQ~\citep{gptq} and GuidedQuant~\citep{GuidedQuant}, which leverage Hessian-based error compensation to mitigate quantization loss. 
QuaRot~\citep{QuaRot} and OstQuant~\citep{OstQuant} leverage rotation matrices to transform the parameter space, thereby improving the distribution of weights and activations across the quantization domain. 
MQuant~\citep{MQuant} and MBQ~\citep{mbq} enhance multimodal PTQ by addressing modality disparities and outliers through structured techniques and gradient-based balancing, respectively, yielding improved accuracy and efficiency.
While these methods are hardware-friendly and straightforward to implement, the quantization error grows sharply below 4-bit, limiting practicality at ultra-low bit allocation.

In contrast to SQ, \textbf{Vector Quantization (VQ) }partitions weights into subvectors and approximates them using a codebook of limited prototypes. Compared with SQ, VQ offers stronger representational capacity and better accuracy retention under 3-bit and even lower precision.
PCDVQ~\citep{PCDVQ} decouples magnitude and direction in polar coordinates and uses distribution-aligned codebooks, delivering strong performance even at 2-bit precision. 
VPTQ ~\citep{vptq} employs channel-wise second-order optimization, efficient codebook initialization, and residual/outlier handling to achieve ultra-low-bit quantization, improving accuracy while reducing calibration time and boosting inference throughput. QuIP\#~\citep{QuIP4PLUS} achieves state-of-the-art extreme compression by integrating structured transforms, lattice codebooks, and lightweight fine-tuning, enabling 3-bit models to outperform 4-bit baselines.

A comprehensive investigation of vector quantization (VQ) for VLMs remains absent, and a general-purpose framework has yet to be established. Two fundamental challenges underpin this gap. First, modality-induced weight heterogeneity poses a significant hurdle, visual and textual tokens exhibit markedly different statistical properties, resulting in mixed-distribution weights that cannot be effectively represented by a unified codebook. Second, neglecting first-order gradients leads to error compensation mismatch, prevailing methods, such as GPTQ~\citep{gptq} and YAQA~\citep{YAQA}, operate under the assumption of near-zero gradients and consequently restrict compensation to second-order, Hessian-based approximations, thereby leaving first-order contributions unaccounted for. Some first-order enhanced methods, such as FOEM~\citep{FOEM}, still retain a GPTQ-style row-wise formulation, modeling curvature per output channel rather than jointly across input and output channels at the layer level.

To tackle these gaps and challenges, we introduce MGVQ. The method integrates multi-dimensional sensitivity analysis, structured mixed-precision allocation, and gradient-aware compensation, and is specifically designed to optimize ultra-low-bit quantization for VLMs.

%% file: sec_4.preliminaries.tex
\section{Preliminaries}\label{sec_Preliminaries}
\begin{figure}[t]
  \centering
  \includegraphics[width=0.9\linewidth]{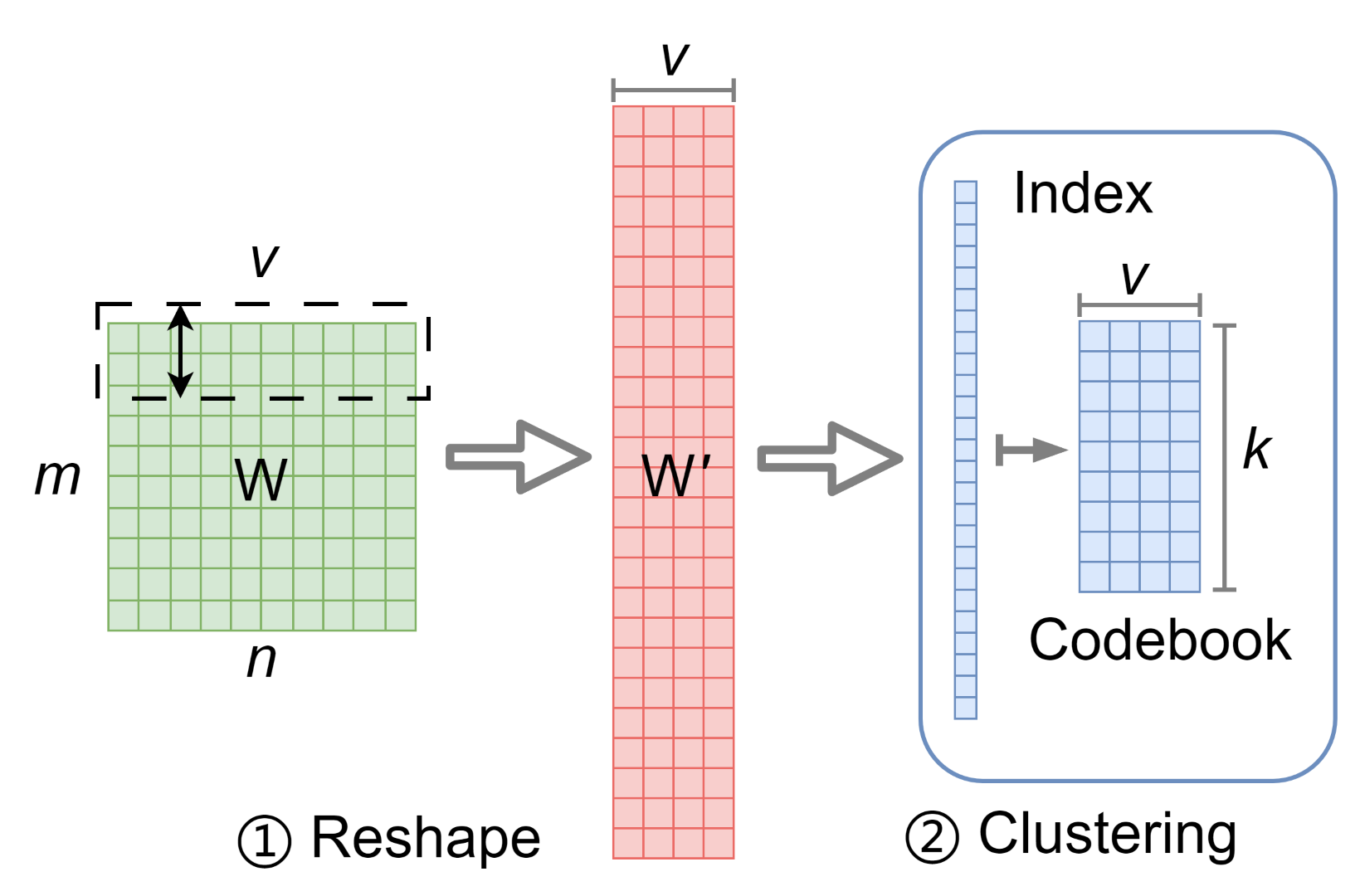}
  \caption{Vector Quantization.}
  %\vspace{-4mm}
  \label{img4_Vector_Quantization}
\end{figure}
\textbf{Vector Quantization in Post-Training Quantization.} In ultra-low-bit PTQ, VQ has attracted increasing attention due to its superior ability to model weight distributions and achieve higher compression ratios compared with scalar quantization. The core idea of VQ is to jointly encode correlated dimensions within small subspaces, approximating each weight sub-vector with a finite set of codewords. Formally, consider a weight matrix $W\in\mathbb{R}^{m\times n}$. Given a block size $v$ (with zero-padding applied if $v\nmid m\,n$), the matrix is reshaped into:
\begin{equation}
W'\in\mathbb{R}^{M\times v}, \qquad M=\frac{m\,n}{v},
\end{equation} 
where the $i$-th row $W'_i\in\mathbb{R}^{1\times v}$ corresponds to a weight vector block of length 
$v$.

A codebook $C\in\mathbb{R}^{K\times v}$ of size $K=2^{n}$ is then constructed, where $n$ denotes the index bitwidth. Each vector block is quantized by selecting its nearest codeword from the codebook under Euclidean distance (noting that the Frobenius norm reduces to $\ell_2$ distance in the vector case):
\begin{equation}\label{eq:VQ}
\begin{aligned}
\mathrm{VQ}(W') =
\Bigl\{\, j_i \;\big|\;&
j_i = \argmin_{j \in \{1,\dots,K\}}
\bigl\| W'_i - C_j \bigr\|_2^2, \\
& i = 1,\dots,M
\Bigr\}.
\end{aligned}
\end{equation}
Finally, the quantized weight matrix $\hat W$ is reconstructed by replacing each block with its assigned codeword and reshaping back to the original dimensions:
\begin{equation}
\hat W=\mathrm{reshape}(\hat W',m,n).
\end{equation} 
This formulation highlights how VQ leverages clustering in a shared codebook to exploit structural redundancy, thereby retaining stronger representational capacity under ultra-low-bit settings compared to scalar quantization.

%When applying compression techniques like quantization, explicitly assessing the global impact of weight perturbations is essential. A straightforward approach is to directly measure changes in task-specific loss. Nevertheless, such an approach is often unreliable, as it depends heavily on the chosen evaluation dataset and cannot be easily decomposed across layers or parameters. 

%% file: sec_5.method.tex
\section{Method}\label{sec_Method}
% \vspace{-3mm}
The proposed MGVQ (Synergizing Multi-dimensional Sensitivity-Aware and Gradient-Hessian Fusion for Vector Quantization) framework is built on the coordinated design of two key modules: channel-sensitivity-driven structured mixed-precision quantization and gradient-aware error compensation.
To address multimodal inputs (image tokens and text tokens), MGVQ first evaluates the sensitivity of input and output channels from both global sensitivity and local functional contribution, and fuses these metrics to form the basis for quantization resource allocation. The weight matrix is then reordered and partitioned into 
$2\times2$ structured sub-blocks, followed by closed-form bit allocation under a global bit budget, ensuring that highly sensitive regions receive higher bitwidth. Finally, MGVQ incorporates both first-order gradient and second-order Hessian information to refine quantization results through fine-grained error compensation, achieving a balanced trade-off between model accuracy and compression efficiency.
The detailed algorithmic procedures of the two modules are provided in Appendix~\ref{appendix:algo}.
%\vspace{-8mm}
\subsection{Sensitivity-driven structured mixed-precision quantization (SSMQ)}\label{method1}
\begin{figure*}[t]
    \centering
    %\vspace{-10mm}
    \includegraphics[width=1\linewidth]{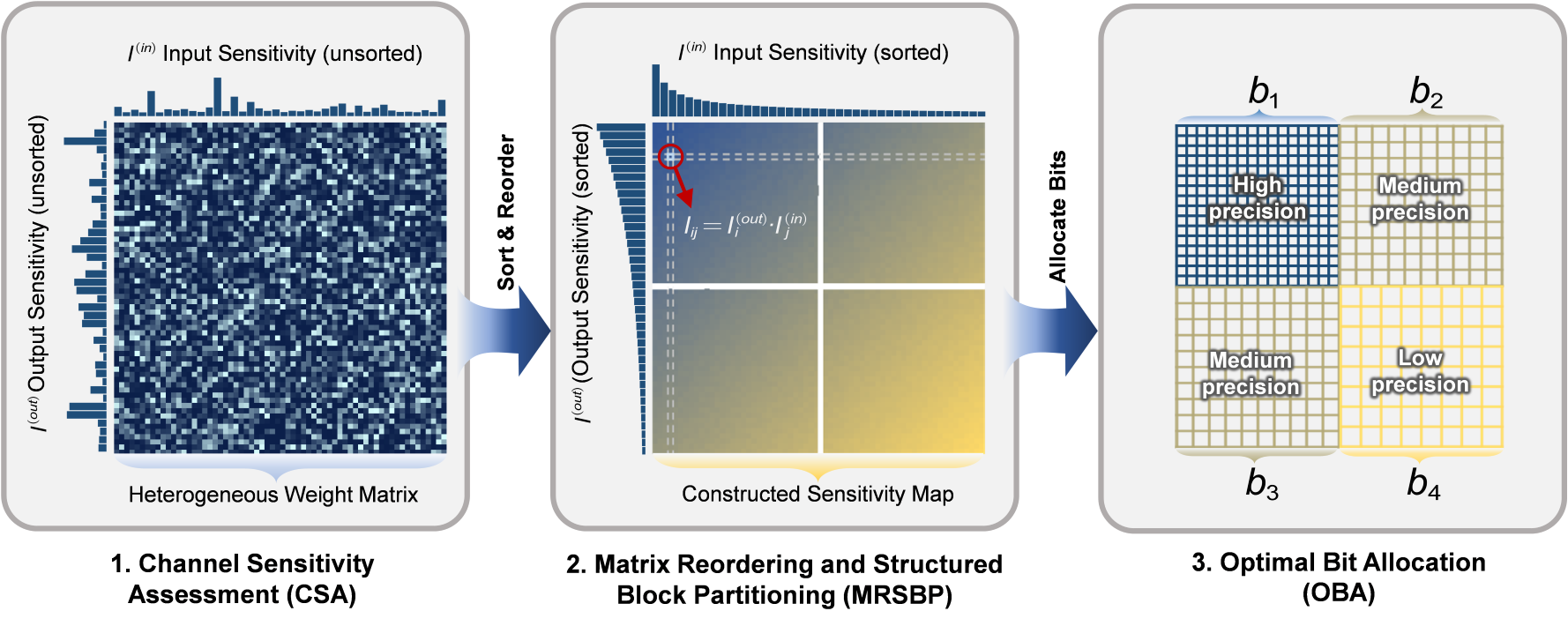}
    % \vspace{-10mm}
    \caption{Overview of sensitivity-driven structured
mixed-precision quantization (SSMQ) }
    \label{img5_CSMQ}
   % \vspace{-4mm}
\end{figure*}

The primary focus in mixed-precision quantization lies in allocating limited bit budgets effectively, so that critical parameters receive finer precision. To this end, we design a channel-sensitivity-driven structured quantization framework, as observed in Figure \ref{img5_CSMQ}, which proceeds in three steps: channel sensitivity assessment (CSA), matrix reordering and structured block partitioning (MRSBP), and optimal bit allocation (OBA).

\textbf{Step 1: Channel Sensitivity Assessment (CSA)}

When compressing models, it is essential to assess the global impact of perturbations. A straightforward approach is to directly measure changes in task-specific loss. Nevertheless, such an approach is often unreliable, as it depends heavily on the evaluation dataset and cannot be easily decomposed across layers. To establish a robust and decomposable metric, we construct channel sensitivity by integrating both global and local perspectives.

\paragraph{Global sensitivity.} For a weight matrix $W \in \mathbb{R}^{m \times n}$ (with m output channels and n input channels), we adopt the Hessian of the KL divergence (equivalent to the Fisher Information Matrix) as a measure of global sensitivity (see Appendix~\ref{subsec_Global_Sensitivity_via_KL} for the derivation). It is approximated by a Kronecker factorization:
\begin{equation}
    H \approx H_O \otimes H_I, H_O \in \mathbb{R}^{m \times m}, H_I \in \mathbb{R}^{n \times n},
\end{equation} 
where $H_O$ denotes the output-side Hessian capturing sensitivity along output channels, and $H_I$ denotes the input-side Hessian capturing sensitivity along input channels.

In practice, these components can be estimated from gradients at the sequence level:
\begin{equation}\label{H_I,H_O}
\begin{aligned}
H_I = \mathbb{E}\left[ (\nabla_{W} \ell)^T (\nabla_{W} \ell) \right],\\ H_O = \mathbb{E}\left[ (\nabla_{W} \ell) (\nabla_{W} \ell)^T \right],
\end{aligned}
\end{equation}

where, $\nabla_{W} \ell$ denotes the gradient of the loss $\ell$ with respect to the weights $W$. The global sensitivity of input channel $i$ is defined as the $i$-th diagonal element of $H_I$, and the global sensitivity of output channel $j$ is defined as the $j$-th diagonal element of $H_O$. We denote these quantities as:
\begin{equation}\label{I_g}
I_g^{(in)}[i]=H_I[i], I_g^{(out)}[j]=H_O[j].
\end{equation} 

%\vspace{-3mm}
\paragraph{Local Sensitivity.} To measure the extent of local output influence across weight channels in practical scenarios, we compute the corresponding norm value of output activations as local sensitivity. The local sensitivity of input-output reflects the output energy generated by activation $x$ and weight $W$ across corresponding channels:
%\vspace{-5mm}
\begin{equation}\label{I_l}
\begin{aligned}
    I_l^{(in)}[i] = \mathbb{E} \left[ \| x\cdot W_{:,i} \|_2^2 \right],\\ I_l^{(out)}[j] = \mathbb{E} \left[ \| x \cdot W_{j,:} \|_2^2 \right].
\end{aligned}
\end{equation}
%\vspace{-12mm}

%\vspace{-5mm}
\paragraph{Combined Sensitivity.} The normalized global and local indicators are fused to obtain the final sensitivity scores:
\begin{equation}\label{I}
\begin{aligned}
I^{(in)}[i]  &= \log\!\left( \hat{I}_g^{(in)}[i] \cdot \hat{I}_l^{(in)}[i] \right), \\
I^{(out)}[j] &= \log\!\left( \hat{I}_g^{(out)}[j] \cdot \hat{I}_l^{(out)}[j] \right),
\end{aligned}
\end{equation}
where $\hat{\cdot}$ denotes min–max normalization. This fusion balances global and local sensitivity, enabling an accurate assessment of quantization difficulty. Consequently, it provides a reliable basis for bit allocation.

\textbf{Step 2: Matrix Reordering and Structured Block Partitioning (MRSBP)}

To cluster parameters with similar sensitivity, we sort input and output channels in descending order of $I^{(in)}$ and $I^{(out)}$. The reordered weight matrix $W'$ is then partitioned into $2\times2$ structured blocks. The sensitivity of each element $(i,j)$ is defined as the product of the corresponding input and output channel sensitivity:
\begin{equation}\label{I_i_j}
I_{i,j} = I^{(out)}[i] \cdot I^{(in)}[j].
\end{equation} 
Given partitioning cut points along the input$/$output dimensions, the matrix is divided into four sub-blocks $\left\{blk_t\right\}_{t=1}^4$. With a total bit budget B, the goal is to allocate bits $\left\{b_t\right\}$ to these sub-blocks to maximize the sensitivity–bit efficiency ratio:

\begin{equation}\label{sensitivity_bit_efficiency_ratio}
\max_{\{b_t\}} \sum_{t=1}^4 \frac{S_t}{b_t}, \quad \text{s.t.} \ \sum_{t=1}^4 b_t = B, \ b_t > 0,
\end{equation} 
where $S_t = \sum_{{(i,j) \in {blk}_t}} I_{i,j}$ is the aggregated sensitivity of block $t$. This ensures that more sensitive regions receive finer quantization.

\textbf{Step 3: Optimal Bit Allocation (OBA)}

Applying Lagrangian optimization yields the closed-form solution for the optimal bit allocation:
\begin{equation}\label{b_t}
b_t = \frac{B \cdot \sqrt{S_t}}{\sum_{s=1}^4 \sqrt{S_s}}.
\end{equation} 
This solution achieves the global optimum of Eq.\ref{sensitivity_bit_efficiency_ratio} in theory, ensuring the maximization of sensitivity return per bit of resource.The detailed derivation procedure is provided in Appendix~\ref{appendix:bit_allocation}.

\subsection{Gradient-Aware Error Compensation (GAEC)}

To bridge the gap between quantization error minimization and task loss minimization, we introduce GAEC. It models the gradient drift caused by quantization~\citep{FOEM} and integrates it into a curvature-based compensation framework~\citep{YAQA}, yielding compensation directions that better match the task loss.

\textbf{Step 1: Formulating the Joint First- and Second-Order Optimization Objective}

We frame the quantization problem as minimizing the task loss induced by the quantization residual. We define the residual $E$ between the original floating-point weights $W$ and their quantized counterparts $\hat{W}$ as the core optimization variable:
\begin{equation}\label{E}
    E \triangleq W - \hat{W}.
\end{equation}
Optimizing $\hat{W}$ to minimize loss is mathematically equivalent to optimizing $E$. To render the high-dimensional Hessian computationally tractable, we adopt a Kronecker-factored approximation ($H \approx H_{O} \otimes H_{I}$). This structural simplification allows us to formulate the joint optimization objective as a quadratic form balancing first-order correction and second-order regularization:
\begin{equation}\label{min(E)}
\min_{\hat{W}} \underbrace{ \mathrm{vec}(E)^T\nabla\mathcal{L}}_{\text{First-order}} + \underbrace{\frac{1}{2} \mathrm{vec}(E)^T(H_{O}\otimes H_{I})\text{vec}(E)}_{\text{Second-order}}.
\end{equation}
Here, $H_{O}$ and $H_{I}$ capture the output and input channel sensitivities, respectively. Conventional second-order methods typically assume a local optimum ($\nabla\mathcal{L}\approx 0$), but accumulated quantization errors shift the weights away from the original optimum, inducing non-negligible first-order gradients. Neglecting this drift, or approximating it with an isotropic proxy ($\nabla\mathcal{L}\approx \beta E$), yields compensation directions that are inconsistent with the curvature structure. To address this, at iteration $t$, we define the current residual as $E^{(t)} = W-\hat W^{(t)}$and explicitly approximate the gradient as the residual projected through the curvature:
\begin{equation}\label{GradientL}
\nabla\mathcal{L}^{(t)} \approx \beta (H_{O} \otimes H_{I}) \text{vec}(E^{(t)}), \quad \beta >0,
\end{equation}
where $\beta$ controls the strength of the curvature-projected gradient correction. Substituting Eq.\ref{GradientL} into the first-order term of Eq.\ref{min(E)} yields the following convex quadratic subproblem in $\text{vec}(E)$:
%\begin{equation}\label{min(E)_joint}
%\min_{E} \frac{1}{2} \langle E, (1 + 2\beta)(H_{O}\otimes H_{I})[E] \rangle.
%\end{equation}
\begin{equation}\label{min(E)_joint}
\begin{aligned}
\min_{\hat{W}} \beta \cdot \text{vec}(E)^T(H_{O}\otimes H_{I})\text{vec}(E^{(t)})+ \\+ \frac{1}{2} \cdot \text{vec}(E)^T(H_{O}\otimes H_{I})\text{vec}(E)
\end{aligned}
\end{equation}
This formulation yields a curvature-aware and gradient-drift-aware compensation objective. It preserves the Kronecker structure for efficient block-wise solving while explicitly accounting for the non-negligible first-order effects induced by quantization.

\textbf{Step 2: Deriving the Curvature-Aligned Closed-Form Solution}

To simplify optimization, we exploit the structure of the Hessian blocks via their block-LDL decompositions:
\begin{equation}
\begin{aligned}
    H_O = (L_O + I) D_O (L_O + I)^T, \\  H_I = (L_I + I) D_I (L_I + I)^T,
\end{aligned}
\end{equation}
where \( L_O, L_I \) are lower triangular matrices with zero diagonals, \( D_O, D_I \) are diagonal matrices of positive entries, and \( I \) is the identity matrix. We define $Z$ and $A^{(t)}$: 
\begin{equation}\label{A}
\begin{aligned}
    Z \;\triangleq\; (L_O + I)^T E \, (L_I + I), \\ A^{(t)}\triangleq(L_O+I)^{-T}E^{(t)}(L_I+I)^{-1}.
\end{aligned}
\end{equation}
Then, Eq.\ref{min(E)_joint} can be decoupled element-wise into
\begin{equation}\label{eq:new_min_taget}
    \min_{Z}\ \sum_{i,j}\Big[\tfrac12\lambda_{ij} Z_{ij}^2+\beta\lambda_{ij} A^{(t)}_{ij} Z_{ij}\Big], 
\end{equation}
where the curvature sensitivity is defined as:
\begin{equation}
\lambda_{ij} = D_O(i) D_I(j) > 0.
\end{equation}
Differentiating Eq.\ref{eq:new_min_taget} with respect to $Z_{ij}$, the sensitivity term $\lambda_{ij}$ acts on both the quadratic and linear terms and thus cancels out in the optimality condition. This yields a remarkably simple closed-form solution:
\begin{equation}
Z_{ij}^{*} = -\beta A_{ij}^{(t)}.
\end{equation}
For the subsequent update step, we explicitly define the correction target $T$ based on this optimal solution:
\begin{equation}\label{T}
T \triangleq \beta A^{(t)}.
\end{equation}
\textbf{Step 3: Fixed-Point Update Under Quantization Constraints (Projection)}

We integrate second-order feedback and the above first-order correction into a single projection step. Let $E^{(t)}=W-\hat{W}^{(t)}$. Define the target
\begin{equation}\label{eta}
\eta \triangleq W + L_O^T E^{(t)} L_I + L_O^T E^{(t)} + E^{(t)} L_I - T,
\end{equation}
and update by projection onto $\mathcal{Q}$:
\begin{equation}\label{w_hat_new}
\hat{W}^{(t+1)} \leftarrow \mathcal{Q}(\eta).
\end{equation}
By iteratively solving Eq.\ref{w_hat_new}, GAEC efficiently finds the optimal quantized weights that balance both the immediate quantization error and the task-loss gradient drift. Notably, as $\beta \to 0$, the method reduces to the purely second-order update without the first-order correction.

%% file: sec_6.experiments.tex
\section{Experiments}
\subsection{Experimental Settings}
%\vspace{-2mm}
\input{tab_main_res}
\input{tab_ablation1}
\paragraph{Calibration set and evaluation benchmarks.} To collect the statistics required for quantization sensitivity analysis, we adopt the improved COCO image-caption dataset provided by ShareGPT4V~\citep{sharegpt4v}. A random subset of 128 image–text pairs is selected as the calibration set, which is used to compute Hessian information and channel importance scores. Model performance is evaluated with the LMMs-Eval benchmark suite~\citep{llms_eval} across a wide range of vision–language tasks, including:
(1) Text recognition and understanding: OCRBench~\citep{ocrbench} (scene text recognition), TextVQA\citep{textvqa} (text-centric visual question answering).
(2) Visual perception: VizWiz~\citep{vizwiz} (QA on everyday photos designed for visually impaired users), SEED-Bench~\citep{seed-bench} (a multimodal benchmark for generation and understanding).
(3) Visual reasoning: ScienceQA~\citep{scienceqa} (science QA with multimodal inputs), MMMU~\cite{mmmu} (multi-discipline multimodal understanding and reasoning).
% \vspace{-4mm}

\paragraph{Models and quantization settings.}
We select three representative families of VLMs with both small and large versions to evaluate the generality of MGVQ:
LLaVA-onevision~\citep{llava}(parameter sizes of 7B and 72B, with the VLM backbone based on Qwen2-7B/72B and the vision encoder SigLIP-400M~\citep{siglip}).
InternVL2~\cite{internvl}(parameter sizes of 8B and 26B, with the VLM backbone InternLM2-8B/20B and the vision encoder InternViT-300M/6B).
Qwen2-VL~\citep{Qwen2-VL}(parameter sizes of 7B and 72B, with VLM backbone Qwen2-7B/72B and a vision encoder of ~675M parameters).
We evaluate 3-bit and 2-bit configurations for each model, comparing MGVQ against strong PTQ baselines. Scalar quantization (SQ) baselines include RTN (uniform quantization), GPTQ ~\citep{gptq}(Hessian-guided), AWQ~\citep{awq} (outlier-aware), and MBQ~\citep{mbq} (recent mixed-precision method). Vector quantization (VQ) baselines include VPTQ~\citep{vptq} and QuIP\#~\cite{QuIP4PLUS}, the latter being state-of-the-art for LLMs.
% \vspace{-4mm}

\paragraph{Implementation details.}
MGVQ first performs row–column reordering of weight matrices, partitions them into four blocks, and applies vector quantization separately to each block. The vector length is set to 4. K-means clustering is initialized with k-means++ and run for 100 iterations. All experiments are conducted on NVIDIA RTX A6000 GPUs.

\subsection{Main Results}
% \vspace{-2mm}
% \input{tab_main_res}
% \input{tab/main_res1}
% \input{tab/main_res2}
% \input{tab/main_res3}

According to Table~\ref{tab:main}, we report the average accuracy across the six datasets introduced above. MGVQ consistently achieves the best overall accuracy under both 3-bit and 2-bit quantization settings. 
In 3-bit quantization, MGVQ outperforms the best existing baselines by 0.2--1.5 percentage points on average. For example, on Qwen2-VL-72B, MGVQ achieves 77.8\%, slightly higher than MBQ (77.6\%) and VPTQ (77.5\%). 
In 2-bit quantization, the advantage of MGVQ is more pronounced, improving over the strongest baseline by 1.3--4.9 percentage points. On InternVL2-26B, MGVQ reaches 71.4\%, compared to QuIP\# (67.0\%) and MBQ (66.5\%). 
MGVQ also significantly narrows the gap between quantized and full-precision models. 
For instance, in the LLaVA-onevision-7B 2-bit case, the FP16 model achieves 66.9\%, while MGVQ reaches 61.5\% (only $-5.4$). By contrast, the best baseline (VPTQ/MBQ) scores 59.6\% ($-7.3$). This demonstrates MGVQ's effectiveness in mitigating quantization-induced accuracy loss.
In addition, end-to-end inference efficiency results are reported in Appendix~\ref{appendix:speed}, which show consistent speedups across both prefilling and decoding stages.
Detailed per-dataset results for each model are provided in Appendix~\ref{appendix:all_res}.

\subsection{Ablation Studies}
%\input{tab_ablation1}
% \vspace{-3mm}
We conduct ablations on LLaVA-onevision-7B~\citep{llava} and Qwen2-VL-7B~\citep{Qwen2-VL}, focusing on 2-bit and 3-bit scenarios across text recognition, visual perception, and visual reasoning tasks. We study the necessity and contributions of two core modules:
SSMQ (sensitivity-driven structured mixed-precision quantization ), and GAEC (gradient-aware error compensation).
Baselines include vanilla VQ (K-means), VPTQ, and GPTVQ.
%\vspace{-3mm}
\paragraph{Joint effectiveness of SSMQ and GAEC.} As shown in Table~\ref{tab:ablation1}, both modules are necessary and complementary.
On LLaVA-onevision-7B (2-bit), without SSMQ/GAEC the average accuracy is only 53.3\%. Adding SSMQ improves it to 60.4\%, while GAEC alone yields 56.8\%. Combining both achieves 61.5\%, narrowing the FP16 gap to 5.4 and outperforming single-module gains by +1.2 and +4.7, respectively.
On Qwen2-VL-7B (3-bit), the average score rises from 64.5\% (no modules) to 71.9\% (both modules), again showing strong synergy.
%\vspace{-6mm}
\paragraph{Effectiveness of SSMQ.} Table~\ref{tab:ablation2} compares SSMQ with existing VQ methods. On LLaVA-onevision-7B (2-bit), vanilla VQ achieves 53.3\%, VPTQ scores 59.6\%, while SSMQ reaches 60.4\%. Notably, in ScienceQA, accuracy improves from 73.1\% (K-means) to 82.1\%, and in TextVQA from 61.0\% to 67.9\%, validating SSMQ’s advantage in dynamically allocating precision to channels under hybrid-distribution weights in VLMs.
\input{tab_ablation2}

\paragraph{Effectiveness of GAEC.} From Table~\ref{tab:ablation3}, GAEC improves over GPTVQ and VPTQ by explicitly incorporating first-order gradient terms into second-order error compensation. Traditional second-order approaches such as GPTVQ (65.7\%) underestimate small gradient regions (0–0.001), leading to insufficient compensation. GAEC alleviates this by leveraging gradient residuals, achieving 68.2\% on Qwen2-VL-7B (3-bit), surpassing GPTVQ (65.7\%) and VPTQ (67.3\%). Gains are especially evident on gradient-sensitive tasks: OCRBench improves from 67.1\% to 69.6\%, and MMMU from 43.9\% to 45.9\%, demonstrating reduced error accumulation across layers and modalities.
\input{tab_ablation3}
%\vspace{-6mm}

%% file: tab_main_res.tex
% \usepackage{multirow}
\begin{table*}[t]
\centering
\caption{Under the 2 - bit and 3 - bit configurations of MGVQ, a comparison is conducted between it and diverse quantization methods of VLMs.}
\vspace{-2mm}
\label{tab:main}
\resizebox{1.0\textwidth}{!}{
\begin{tabular}{c|c|llllll}
\hline
Bit & Method & \multicolumn{1}{c}{LLaVA-onevision-7B} & LLaVA-onevision-72B & Qwen2-VL-7B & Qwen2-VL-72B & InternVL2-8B & InternVL2-26B \\ \hline
\multicolumn{1}{l|}{FP16} & - & 66.9 & 74.3 & 73.1 & 78.1 & 71.7 & 74.6 \\ \hline
 & RTN & 47.9 & 72.1 & 65.4 & 75.0 & 69.0 & 73.3 \\
 & GPTQ & 63.4 & 72.3 & 67.9 & 76.6 & 67.2 & 72.3 \\
 & AWQ & 60.4 & 55.1 & 70.3 & 77.5 & 69.8 & 73.5 \\
 & MBQ & 64.8 & 73.6 & 70.9 & 77.6 & 70.4 & 73.8 \\
 & VPTQ & 65.3 & 73.6 & 71.3 & 77.5 & 70.6 & 73.9 \\
\multirow{-6}{*}{3} & \cellcolor[HTML]{F2F3F5}MGVQ & \cellcolor[HTML]{F2F3F5}65.8 & \cellcolor[HTML]{F2F3F5}74.0 & \cellcolor[HTML]{F2F3F5}71.9 & \cellcolor[HTML]{F2F3F5}77.8 & \cellcolor[HTML]{F2F3F5}71.1 & \cellcolor[HTML]{F2F3F5}74.2 \\ \hline
 & RTN & 25.5 & 42.2 & 51.8 & 64.0 & 44.3 & 59.4 \\
 & GPTQ & 38 & 50.5 & 52.5 & 67.1 & 46.6 & 60.2 \\
 & AWQ & 40.4 & 42.2 & 53.0 & 66.3 & 48.5 & 61.4 \\
 & MBQ & 59.6 & 66.5 & 65.1 & 71.1 & 59.1 & 66.5 \\
 & VPTQ & 59.6 & 70.2 & 67.3 & 72.6 & 61.3 & 66.3 \\
 & QuIP\# & 58.9 & 70.9 & 68.2 & 73.9 & 62.4 & 67.0 \\
\multirow{-7}{*}{2} & \cellcolor[HTML]{F2F3F5}MGVQ & \cellcolor[HTML]{F2F3F5}61.5 & \cellcolor[HTML]{F2F3F5}72.4 & \cellcolor[HTML]{F2F3F5}70.0 & \cellcolor[HTML]{F2F3F5}75.8 & \cellcolor[HTML]{F2F3F5}66.7 & \cellcolor[HTML]{F2F3F5}71.4 \\ \hline
\end{tabular}
}
\end{table*}

%% file: tab_ablation1.tex
% \usepackage{multirow}

\begin{table*}[!h]
\caption{Ablation experiments on SSMQ and GAEC} 
%\vspace{-2mm}
\label{tab:ablation1}
\resizebox{1.0\textwidth}{!}{
\begin{tabular}{l|c|cc|ccccccc}
\hline
Model & Bit & SSMQ & GAEC & \multicolumn{1}{c}{MMMU} & \multicolumn{1}{c}{SEED} & \multicolumn{1}{c}{OCRBench} & \multicolumn{1}{c}{VizWiz} & \multicolumn{1}{c}{ScienceQA} & \multicolumn{1}{c}{TextVQA} & Average \\ \hline
\multirow{4}{*}{LLaVA-onevision-7B} & \multirow{4}{*}{2} & \ding{55} & \ding{55} & 33.1 & 51.3 & 50.1 & 51.0 & 73.1 & 61.0 & 53.3 \\
 &  & \checkmark & \ding{55} & 38.9 & 65.1 & 52.3 & 55.8 & 82.1 & 67.9 & 60.4 \\
 &  & \ding{55} & \checkmark & 35.2 & 56.9 & 51.4 & 53.9 & 76.9 & 66.3 & 56.8 \\
 &  & \checkmark & \checkmark & 40.1 & 66.3 & 54.5 & 56.1 & 83.1 & 68.9 & 61.5 \\ \hline
\multirow{4}{*}{Qwen2-VL-7B} & \multirow{4}{*}{3} & \ding{55} & \ding{55} & 41.4 & 63.2 & 68.3 & 61.4 & 81.1 & 71.4 & 64.5 \\
 &  & \checkmark & \ding{55} & 47.2 & 68.3 & 73.9 & 66.3 & 83.1 & 77.9 & 69.5 \\
 &  & \ding{55} & \checkmark & 45.9 & 68.9 & 69.6 & 65.6 & 82.2 & 76.9 & 68.2 \\
 &  & \checkmark & \checkmark & 49.3 & 70.5 & 78.1 & 68.1 & 84.3 & 80.9 & 71.9 \\ \hline
\end{tabular}
}
\end{table*}

\begin{comment}
\begin{table}[t]
\caption{Ablation experiments on CSMQ and GSCM.}
\label{tab:ablation1}
\vspace{-2mm}
\centering
\small
\setlength{\tabcolsep}{4pt}
\begin{tabular}{l|cccc|cccc}
\hline
& \multicolumn{4}{c|}{LLaVA-OneVision-7B (2-bit)} 
& \multicolumn{4}{c}{Qwen2-VL-7B (3-bit)} \\
Benchmark 
& 00 & 10 & 01 & 11 
& 00 & 10 & 01 & 11 \\
\hline
MMMU        & 33.1 & 38.9 & 35.2 & 40.1 & 41.4 & 47.2 & 45.9 & 49.3 \\
SEED        & 51.3 & 65.1 & 56.9 & 66.3 & 63.2 & 68.3 & 68.9 & 70.5 \\
OCRBench   & 50.1 & 52.3 & 51.4 & 54.5 & 68.3 & 73.9 & 69.6 & 78.1 \\
VizWiz     & 51.0 & 55.8 & 53.9 & 56.1 & 61.4 & 66.3 & 65.6 & 68.1 \\
ScienceQA  & 73.1 & 82.1 & 76.9 & 83.1 & 81.1 & 83.1 & 82.2 & 84.3 \\
TextVQA    & 61.0 & 67.9 & 66.3 & 68.9 & 71.4 & 77.9 & 76.9 & 80.9 \\
\hline
Average    & 53.3 & 60.4 & 56.8 & 61.5 & 64.5 & 69.5 & 68.2 & 71.9 \\
\hline
\end{tabular}
\end{table}
\end{comment}

%% file: tab_ablation2.tex
% \usepackage{multirow}
\begin{table}[!h]
\caption{Effectiveness of SSMQ on LLaVA-OneVision-7B under 2-bit quantization}

\label{tab:ablation2}
\centering
\small
%\resizebox{1.0\textwidth}{!}{
\begin{tabular}{l|ccc}
\hline
Benchmark & Kmeans & VPTQ & OURS (SSMQ) \\
\hline
MMMU        & 33.1 & 38.7 & 38.9 \\
SEED        & 51.3 & 64.6 & 65.1 \\
OCRBench   & 50.1 & 51.1 & 52.3 \\
VizWiz     & 51.0 & 55.3 & 55.8 \\
ScienceQA  & 73.1 & 80.4 & 82.1 \\
TextVQA    & 61.0 & 67.3 & 67.9 \\
\hline
Average    & 53.3 & 59.6 & 60.4 \\
\hline
\end{tabular}
%}
\end{table}

%% file: tab_ablation3.tex
% \usepackage{multirow}
\begin{comment}
\begin{table}[!h]
\caption{Effectiveness of CSMQ}
\vspace{-2mm}
\label{tab:ablation2}
\resizebox{1.0\textwidth}{!}{
\begin{tabular}{l|l|l|lllllll}
\hline
Model & Bit & Method & \multicolumn{1}{c}{MMMU} & \multicolumn{1}{c}{SEED} & \multicolumn{1}{c}{OCRBench} & \multicolumn{1}{c}{VizWiz} & \multicolumn{1}{c}{ScienceQA} & \multicolumn{1}{c}{TextVQA} & Average \\ \hline
\multirow{3}{*}{LLaVA-onevision-7B} & \multirow{3}{*}{2} & Kmeans & 33.1 & 51.3 & 50.1 & 51.0 & 73.1 & 61.0 & 53.3 \\
 &  & VPTQ & 38.7 & 64.6 & 51.1 & 55.3 & 80.4 & 67.3 & 59.6 \\
 &  & OURS(CSMQ) & 38.9 & 65.1 & 52.3 & 55.8 & 82.1 & 67.9 & 60.4 \\ \hline
\end{tabular}
}
\end{table}
\end{comment}

\begin{table}[!h]
\caption{Effectiveness of GAEC on Qwen2-VL-7B under 2-bit quantization.}
\label{tab:ablation3}
%\vspace{-2mm}
\centering
\small
\begin{tabular}{l|ccc}
\hline
Benchmark & GPTVQ & VPTQ & OURS (GAEC) \\
\hline
MMMU        & 43.9 & 44.9 & 45.9 \\
SEED        & 65.2 & 68.1 & 68.9 \\
OCRBench   & 67.1 & 67.2 & 69.6 \\
VizWiz     & 63.2 & 65.6 & 65.6 \\
ScienceQA  & 80.1 & 81.1 & 82.2 \\
TextVQA    & 74.5 & 76.9 & 76.9 \\
\hline
Average    & 65.7 & 67.3 & 68.2 \\
\hline
\end{tabular}
\end{table}

%% file: sec_7.conclusion.tex
\section{Conclusion}
%\vspace{-4mm}
This work addresses the unique challenges of applying vector quantization to vision-language models (VLMs), where modality-induced weight heterogeneity and the non-negligible role of first-order gradients lead to severe performance degradation under low-bit settings. We propose \textbf{MGVQ}, a multi-dimensional sensitivity-aware vector quantization framework that integrates (1) modality-induced weight heterogeneity, and (2) gradient-aware error compensation. By jointly leveraging global-local sensitivity measures and efficient Kronecker/Block-LDL decomposition, MGVQ achieves fine-grained bit allocation and accurate error correction. 
Extensive experiments across diverse VLM families (LLaVA-onevision, InternVL2, Qwen2-VL) and model scales (7B–72B) demonstrate that MGVQ consistently outperforms existing SQ and VQ baselines, especially in the extreme 2-bit regime. Notably, MGVQ significantly reduces the quantization–FP16 gap, highlighting its effectiveness in mitigating cross-modal error accumulation. Ablation studies further confirm the complementary contributions of sensitivity-driven allocation and gradient-aware compensation.
Our findings suggest that well-founded quantization strategies are crucial for enabling efficient deployment of large-scale multimodal models. Beyond the immediate improvements in VLM quantization, the proposed MGVQ framework offers a general perspective on integrating structural sensitivity analysis and gradient-informed optimization, which may inspire future research on compressing and accelerating multimodal foundation models. In the future, we plan to extend MGVQ to more complex multimodal scenarios, such as video-language understanding and multi-task joint modeling, to further explore its generalization potential.

%% file: sec_9.Impact_Statement.tex
\section{Impact Statement}
This paper presents work whose goal is to advance the field of Machine Learning. There are many potential societal consequences of our work, none which we feel must be specifically highlighted here.

%% file: sec_8.appendix.tex
\clearpage
\appendix
\onecolumn
\section{Appendix}

\subsection{Closed-form Solution for Optimal Bit Allocation}\label{appendix:bit_allocation}
\input{appendix_form_Closed_form_Solution}

%\subsection{Error bound under Kronecker Hessian and first-order gradient approximation}\label{appendix:first_order}
%%\input{appendix_form/Error} 
%\input{appendix_form/Strict Convexity and Well_posedness} 

\subsection{Global Sensitivity via KL Divergence}\label{subsec_Global_Sensitivity_via_KL}
\input{appendix_form_KL_DIVERGENCE}

%\clearpage
%\subsection{Algorithm}\label{appendix:algo}
%\input{pseudo/pseudo_CSMQ1}
%%\input{pseudo/pseudo_GSCM}
%\input{pseudo/pseudo_GSCM1}

\subsection{Additional Results}\label{appendix:speed}
\input{tab_end_to_end_speed}

%\clearpage
\subsection{All Results}\label{appendix:all_res}
\input{tab_llava_res}
\input{tab_intervl}
\input{tab_qwenvl}

\clearpage
\subsection{Algorithm}\label{appendix:algo}
\input{pseudo_pseudo_SSMQ}
\input{pseudo_pseudo_GAEC}

%% file: appendix_form_Closed_form_Solution.tex
The optimal bit allocation $\{b_t\}_{t=1}^4$ that minimizes the objective 

\begin{equation}
\min_{\{b_t\}} \sum_{t=1}^4 \frac{S_t}{b_t}, 
\quad \text{s.t. } \sum_{t=1}^4 b_t = B, \; b_t > 0,
\end{equation}
where $S_t = \sum_{(i,j) \in \text{blk}_t} I_{i,j}$ is the total sensitivity of block $t$, is given by the closed-form expression:
\begin{equation}
b_t = \frac{B \cdot \sqrt{S_t}}{\sum_{s=1}^4 \sqrt{S_s}}, \quad t=1,\ldots,4.
\end{equation}

\textit{Proof}. Define the Lagrangian:
\begin{equation}
\mathcal{L}(b_1,\ldots,b_4,\lambda) 
= \sum_{t=1}^4 \frac{S_t}{b_t} 
+ \lambda \Big( B - \sum_{t=1}^4 b_t \Big).
\end{equation}

Taking the derivative with respect to $b_t$ and setting it to zero gives:
\begin{equation}
\frac{\partial \mathcal{L}}{\partial b_t} 
= -\frac{S_t}{b_t^2} - \lambda = 0 
\;\; \Rightarrow \;\; \frac{S_t}{b_t^2} = -\lambda.
\end{equation}

Hence,
\begin{equation}
\frac{S_1}{b_1^2} = \cdots = \frac{S_4}{b_4^2} = c,
\end{equation}
for some constant $c$, leading to
\begin{equation}
b_t = \sqrt{\tfrac{S_t}{c}}.
\end{equation}

Applying the constraint $\sum_{t=1}^4 b_t = B$, we obtain
\begin{equation}
c = \left( \frac{\sum_{t=1}^4 \sqrt{S_t}}{B} \right)^2.
\end{equation}

Substituting $c$ yields the closed-form solution
\begin{equation}
b_t = \frac{B \cdot \sqrt{S_t}}{\sum_{s=1}^4 \sqrt{S_s}}.
\end{equation}
This ensures that more bits are allocated to blocks with larger sensitivity, achieving globally optimal efficiency.

%% file: appendix_form_KL_DIVERGENCE.tex
In model compression, directly evaluating task loss changes is often unreliable, as it depends on specific datasets and cannot be decomposed across layers. Instead, we adopt the Kullback–Leibler (KL) divergence between the outputs of the full-precision model and the quantized model as a principled measure of quantization sensitivity.

Given a full-precision model $M(W, X)$ with parameters $W$ and its quantized counterpart $M(\hat{W},X)$, the global KL loss is defined as:
\begin{equation}
\mathcal{L}_{\mathrm{KL}}(\hat{W}) = \mathbb{E}_{X\sim\mathcal D} D_{\text{KL}}(M(W, X) \| M(\hat{W}, X))
\end{equation} 
where $X \sim \mathcal{D}$ denotes inputs drawn from the data distribution $D$. For small perturbations around $W$, a second-order Taylor expansion yields:
\begin{equation}
\mathcal{L}_{\mathrm{KL}}(\hat{W}) \approx \frac{1}{2} (\hat{W} - W)^T H_{\mathrm{global}} (\hat{W} - W)
\end{equation} 
where $H_{global}$ is the Hessian of the KL divergence, equivalent to the Fisher Information Matrix:
\begin{equation}
H_{\mathrm{global}} = \mathbb{E}\Big[ \nabla_{W} \ell \; \nabla_{W} \ell^T \Big]
\end{equation} 
Thus, the global Hessian provides a second-order sensitivity metric for weight perturbations, forming the theoretical basis for sensitivity analysis and bit allocation in our quantization framework.

%% file: tab_end_to_end_speed.tex
\begin{comment}
\begin{table}[!h]
\centering
\begin{tabular}{l l c c c}
\toprule
Model & Stage & FP16 (ms) & W3 (ms) & W2 (ms) \\
\midrule
ViT & Prefill (729 tokens) & 11.5 & 10.5 & 9.7 \\
\midrule
\multirow{3}{*}{VLM} 
 & Prefill (512 tokens) & 68.1 & 61.3 & 59.8 \\
 & Prefill (1024 tokens) & 108.6 & 100.5 & 94.7 \\
 & Decode & 29.3 & 18.4 & 25.6 \\
\bottomrule
\end{tabular}
\caption{The end-to-end speed up of LLaVA-onevision-7B on RTX4090 with fused GPU kernels.
Experimental results show that our method achieves approximately $9\%$ (W3) and 15$\%$ (W2) acceleration over FP16 on the Vision Transformer (ViT), delivers an average improvement of about 7–13$\%$ in the VLM Prefill stage, and reaches up to 37$\%$ acceleration in the Decoder stage, demonstrating the inference efficiency of our approach.}
\label{tab:speedup}
\end{table}
\end{comment}

\begin{table}[!h]
\centering
\caption{The end-to-end speed up of LLaVA-onevision-7B on RTX4090 with fused GPU kernels.
Experimental results show that our method achieves approximately $9\%$ (W3) and 15$\%$ (W2) acceleration over FP16 on the Vision Transformer (ViT), delivers an average improvement of about 7–13$\%$ in the VLM Prefill stage, and reaches up to 37$\%$ acceleration in the Decoder stage, demonstrating the inference efficiency of our approach.}
\setlength{\tabcolsep}{4pt}  % 默认是 6pt
\begin{tabular}{l|lccc}
\hline
Model & Stage (tokens) & FP16 (ms) & W3 (ms) & W2 (ms) \\
\hline
ViT & Prefill (729) & 11.5 & 10.5 & 9.7 \\
\hline
VLM & Prefill (512) & 68.1 & 61.3 & 59.8 \\
    & Prefill (1024) & 108.6 & 100.5 & 94.7 \\
    & Decode & 29.3 & 18.4 & 25.6 \\
\hline
\end{tabular}
%\caption{The end-to-end speedup of LLaVA-OneVision-7B on RTX4090 with fused GPU kernels.}
\label{tab:speedup}
\end{table}

%% file: tab_llava_res.tex
% \usepackage{multirow}
\begin{table}[h]
\centering
\caption{Results of LLaVA-onevision-7B. }
\label{tab:llava-7b}
\resizebox{1.0\textwidth}{!}{
\begin{tabular}{c|c|ccccccc}
\hline
Bit & Method & \multicolumn{1}{c}{MMMU} & \multicolumn{1}{c}{SEED} & \multicolumn{1}{c}{OCRBench} & \multicolumn{1}{c}{VizWiz} & \multicolumn{1}{c}{ScienceQA} & \multicolumn{1}{c}{TextVQA} & \multicolumn{1}{c}{Average ($\uparrow$)} \\ \hline
FP16 & - & 46.0 & 71.1 & 62.2 & 60.4 & 85.4 & 76.1 & 66.9 \\ \hline
 & RTN & 34.7 & 10.4 & 35.9 & 59.2 & 86.2 & 60.9 & 47.9 \\
 & GPTQ & 41.9 & 68.7 & 55.7 & 56.4 & 86.4 & 71.3 & 63.4 \\
 & AWQ & 36.6 & 51.5 & 59.3 & 58.5 & 83.2 & 73.0 & 60.4 \\
 & MBQ & 42.0 & 66.4 & 61.1 & 60.7 & 85.0 & 73.3 & 64.8 \\
 & VPTQ & 43.3 & 69.1 & 61.4 & 60.2 & 84.7 & 73.1 & 65.3 \\
\multirow{-6}{*}{3} & \cellcolor[HTML]{F2F3F5}MGVQ & \cellcolor[HTML]{F2F3F5}44.0 & \cellcolor[HTML]{F2F3F5}69.9 & \cellcolor[HTML]{F2F3F5}61.5 & \cellcolor[HTML]{F2F3F5}60.3 & \cellcolor[HTML]{F2F3F5}85.1 & \cellcolor[HTML]{F2F3F5}73.9 & \cellcolor[HTML]{F2F3F5}65.8 \\ \hline
 & RTN & 13.9 & 0.0 & 10.3 & 36.8 & 61.3 & 30.4 & 25.5 \\
 & GPTQ & 30.2 & 8.5 & 29.7 & 50.1 & 70.3 & 39.4 & 38.0 \\
 & AWQ & 30.9 & 9.8 & 35.3 & 50.3 & 71.9 & 43.9 & 40.4 \\
 & MBQ & 37.6 & 63.5 & 52.0 & 56.1 & 81.2 & 67.5 & 59.6 \\
 & VPTQ & 38.7 & 64.6 & 51.1 & 55.3 & 80.4 & 67.3 & 59.6 \\
 & QuIP\# & 37.7 & 62.3 & 51.0 & 55.2 & 80.3 & 67.1 & 58.9 \\
\multirow{-7}{*}{2} & \cellcolor[HTML]{F2F3F5}MGVQ & \cellcolor[HTML]{F2F3F5}40.1 & \cellcolor[HTML]{F2F3F5}66.3 & \cellcolor[HTML]{F2F3F5}54.5 & \cellcolor[HTML]{F2F3F5}56.1 & \cellcolor[HTML]{F2F3F5}83.1 & \cellcolor[HTML]{F2F3F5}68.9 & \cellcolor[HTML]{F2F3F5}61.5 \\ \hline
\end{tabular}
}
\end{table}

% \usepackage{multirow}
\begin{table}[h]
\centering
\caption{Results of LLaVA-onevision-72B.}
\label{tab:llava-72b}
\resizebox{1.0\textwidth}{!}{
\begin{tabular}{c|c|ccccccc}
\hline
Bit & Method & \multicolumn{1}{c}{MMMU} & \multicolumn{1}{c}{SEED} & \multicolumn{1}{c}{OCRBench} & \multicolumn{1}{c}{VizWiz} & \multicolumn{1}{c}{ScienceQA} & \multicolumn{1}{c}{TextVQA} & \multicolumn{1}{c}{Average ($\uparrow$)} \\ \hline
FP16 & - & 56.1 & 78.1 & 73.2 & 69.2 & 90.0 & 79.3 & 74.3 \\ \hline
 & RTN & 53.9 & 77.4 & 68.2 & 66.1 & 89.5 & 77.4 & 72.1 \\
 & GPTQ & 52.7 & 76.0 & 69.7 & 68.3 & 89.3 & 77.9 & 72.3 \\
 & AWQ & 33.4 & 71.2 & 48.7 & 49.3 & 69.2 & 58.8 & 55.1 \\
 & MBQ & 54.4 & 77.6 & 71.6 & 69.0 & 90.3 & 78.5 & 73.6 \\
 & VPTQ & 54.5 & 77.8 & 71.9 & 69.1 & 90.0 & 78.4 & 73.6 \\
\multirow{-6}{*}{3} & \cellcolor[HTML]{F2F3F5}MGVQ & \cellcolor[HTML]{F2F3F5}55.6 & \cellcolor[HTML]{F2F3F5}77.9 & \cellcolor[HTML]{F2F3F5}72.5 & \cellcolor[HTML]{F2F3F5}69.0 & \cellcolor[HTML]{F2F3F5}90.1 & \cellcolor[HTML]{F2F3F5}79.0 & \cellcolor[HTML]{F2F3F5}74.0 \\ \hline
 & RTN & 34.5 & 18.5 & 34.5 & 50.1 & 71.1 & 44.4 & 42.2 \\
 & GPTQ & 47.2 & 30.1 & 40.3 & 58.4 & 74.0 & 52.9 & 50.5 \\
 & AWQ & 33.0 & 17.9 & 31.2 & 54.9 & 69.2 & 47.1 & 42.2 \\
 & MBQ & 48.1 & 70.4 & 67.1 & 60.2 & 83.8 & 69.1 & 66.5 \\
 & VPTQ & 51.3 & 74.6 & 69.0 & 66.3 & 86.8 & 72.9 & 70.2 \\
 & QuIP\# & 52.5 & 75.3 & 69.9 & 66.5 & 86.8 & 74.6 & 70.9 \\
\multirow{-7}{*}{2} & \cellcolor[HTML]{F2F3F5}MGVQ & \cellcolor[HTML]{F2F3F5}53.4 & \cellcolor[HTML]{F2F3F5}75.8 & \cellcolor[HTML]{F2F3F5}71.7 & \cellcolor[HTML]{F2F3F5}68.1 & \cellcolor[HTML]{F2F3F5}87.9 & \cellcolor[HTML]{F2F3F5}77.4 & \cellcolor[HTML]{F2F3F5}72.4 \\ \hline
\end{tabular}
}
\end{table}

%% file: tab_intervl.tex
% \usepackage{multirow}
\begin{table}[h]
\centering
\caption{Results of InternVL2-8B. }
\label{tab:InternVL2-8B}
\resizebox{1.0\textwidth}{!}{
\begin{tabular}{c|c|ccccccc}
\hline
Bit & Method & \multicolumn{1}{c}{MMMU} & \multicolumn{1}{c}{SEED} & \multicolumn{1}{c}{OCRBench} & \multicolumn{1}{c}{VizWiz} & \multicolumn{1}{c}{ScienceQA} & \multicolumn{1}{c}{TextVQA} & \multicolumn{1}{c}{Average ($\uparrow$)} \\ \hline
FP16 & - & 48.0 & 71.6 & 76.5 & 61.1 & 96.2 & 77.0 & 71.7 \\ \hline
 & RTN & 43.7 & 70.3 & 74.0 & 56.0 & 95.6 & 74.6 & 69.0 \\
 & GPTQ & 41.7 & 68.9 & 70.2 & 59.9 & 89.5 & 73.1 & 67.2 \\
 & AWQ & 44.8 & 70.4 & 74.7 & 58.9 & 95.5 & 74.2 & 69.8 \\
 & MBQ & 46.9 & 70.8 & 75.1 & 58.7 & 95.6 & 75.1 & 70.4 \\
 & VPTQ & 47.1 & 70.9 & 75.4 & 59.1 & 95.5 & 75.8 & 70.6 \\
\multirow{-6}{*}{3} & \cellcolor[HTML]{F2F3F5}MGVQ & \cellcolor[HTML]{F2F3F5}47.6 & \cellcolor[HTML]{F2F3F5}71.3 & \cellcolor[HTML]{F2F3F5}75.9 & \cellcolor[HTML]{F2F3F5}59.5 & \cellcolor[HTML]{F2F3F5}95.6 & \cellcolor[HTML]{F2F3F5}76.5 & \cellcolor[HTML]{F2F3F5}71.1 \\ \hline
 & RTN & 33.5 & 10.2 & 34.1 & 50.9 & 72.2 & 65.1 & 44.3 \\
 & GPTQ & 30.4 & 18.9 & 37.9 & 48.1 & 77.9 & 66.3 & 46.6 \\
 & AWQ & 34.5 & 20.7 & 38.2 & 53.2 & 75.8 & 68.8 & 48.5 \\
 & MBQ & 40.3 & 65.8 & 50.4 & 53.3 & 77.3 & 67.3 & 59.1 \\
 & VPTQ & 44.9 & 64.9 & 57.3 & 55.2 & 77.1 & 68.1 & 61.3 \\
 & QuIP\# & 45.3 & 67.3 & 61.2 & 54.1 & 78.2 & 68.4 & 62.4 \\
\multirow{-7}{*}{2} & \cellcolor[HTML]{F2F3F5}MGVQ & \cellcolor[HTML]{F2F3F5}46.2 & \cellcolor[HTML]{F2F3F5}69.3 & \cellcolor[HTML]{F2F3F5}68.4 & \cellcolor[HTML]{F2F3F5}57.6 & \cellcolor[HTML]{F2F3F5}86.4 & \cellcolor[HTML]{F2F3F5}72.3 & \cellcolor[HTML]{F2F3F5}66.7 \\ \hline
\end{tabular}
}
\end{table}

% \usepackage{multirow}
\begin{table}[h]
\centering
\caption{Results of InternVL2-26B. }
\label{tab:InternVL2-26B}
\resizebox{1.0\textwidth}{!}{
\begin{tabular}{c|c|ccccccc}
\hline
Bit & Method & \multicolumn{1}{c}{MMMU} & \multicolumn{1}{c}{SEED} & \multicolumn{1}{c}{OCRBench} & \multicolumn{1}{c}{VizWiz} & \multicolumn{1}{c}{ScienceQA} & \multicolumn{1}{c}{TextVQA} & \multicolumn{1}{c}{Average ($\uparrow$)} \\ \hline
FP16 & - & 47.1 & 76.8 & 77.9 & 66.2 & 97.5 & 82.1 & 74.6 \\ \hline
 & RTN & 46.6 & 75.7 & 75.9 & 64.7 & 96.4 & 80.6 & 73.3 \\
 & GPTQ & 44.8 & 75.8 & 76.0 & 60.9 & 96.3 & 80.1 & 72.3 \\
 & AWQ & 46.4 & 76.2 & 76.4 & 64.5 & 96.7 & 81.0 & 73.5 \\
 & MBQ & 47.1 & 76.3 & 76.5 & 64.5 & 97.3 & 81.1 & 73.8 \\
 & VPTQ & 47.3 & 76.0 & 76.9 & 65.2 & 97.1 & 81.0 & 73.9 \\
\multirow{-6}{*}{3} & \cellcolor[HTML]{F2F3F5}MGVQ & \cellcolor[HTML]{F2F3F5}47.1 & \cellcolor[HTML]{F2F3F5}76.4 & \cellcolor[HTML]{F2F3F5}77.3 & \cellcolor[HTML]{F2F3F5}65.0 & \cellcolor[HTML]{F2F3F5}97.3 & \cellcolor[HTML]{F2F3F5}81.8 & \cellcolor[HTML]{F2F3F5}74.2 \\ \hline
 & RTN & 36.2 & 55.6 & 63.5 & 56.1 & 73.4 & 71.4 & 59.4 \\
 & GPTQ & 37.4 & 58.2 & 64.3 & 55.1 & 72.4 & 73.8 & 60.2 \\
 & AWQ & 38.4 & 55.4 & 64.9 & 57.4 & 74.9 & 77.4 & 61.4 \\
 & MBQ & 43.2 & 65.3 & 70.2 & 58.4 & 83.9 & 78.1 & 66.5 \\
 & VPTQ & 42.4 & 67.4 & 72.3 & 55.1 & 80.3 & 80.1 & 66.3 \\
 & QuIP\# & 44.2 & 69.3 & 71.2 & 56.1 & 80.3 & 81.1 & 67.0 \\
\multirow{-7}{*}{2} & \cellcolor[HTML]{F2F3F5}MGVQ & \cellcolor[HTML]{F2F3F5}46.1 & \cellcolor[HTML]{F2F3F5}73.9 & \cellcolor[HTML]{F2F3F5}73.2 & \cellcolor[HTML]{F2F3F5}60.3 & \cellcolor[HTML]{F2F3F5}93.6 & \cellcolor[HTML]{F2F3F5}81.1 & \cellcolor[HTML]{F2F3F5}71.4 \\ \hline
\end{tabular}
}
\end{table}

%% file: tab_qwenvl.tex
% Qwen2-VL-7B
\begin{table}[h]
\centering
\caption{Results of Qwen2-VL-7B. }
\label{tab:qwen2-VL-7B}
\resizebox{1.0\textwidth}{!}{
\begin{tabular}{c|c|lllllll}
\hline
Bit & Method & \multicolumn{1}{c}{MMMU} & \multicolumn{1}{c}{SEED} & \multicolumn{1}{c}{OCRBench} & \multicolumn{1}{c}{VizWiz} & \multicolumn{1}{c}{ScienceQA} & \multicolumn{1}{c}{TextVQA} & \multicolumn{1}{c}{Average ($\uparrow$)} \\ \hline
FP16 & - & 50.6 & 71.9 & 80.7 & 68.3 & 85.1 & 82.0 & 73.1 \\ \hline
 & RTN & 44.9 & 69.8 & 60.0 & 65.2 & 81.5 & 71.2 & 65.4 \\
 & GPTQ & 43.1 & 68.9 & 74.8 & 64.3 & 79.7 & 76.7 & 67.9 \\
 & AWQ & 44.7 & 70.4 & 76.9 & 68.0 & 82.5 & 79.5 & 70.3 \\
 & MBQ & 47.9 & 70.2 & 76.8 & 67.7 & 82.8 & 79.9 & 70.9 \\
 & VPTQ & 48.3 & 70.5 & 77.6 & 67.9 & 83.4 & 79.9 & 71.3 \\
\multirow{-6}{*}{3} & \cellcolor[HTML]{F2F3F5}MGVQ & \cellcolor[HTML]{F2F3F5}49.3 & \cellcolor[HTML]{F2F3F5}70.5 & \cellcolor[HTML]{F2F3F5}78.1 & \cellcolor[HTML]{F2F3F5}68.1 & \cellcolor[HTML]{F2F3F5}84.3 & \cellcolor[HTML]{F2F3F5}80.9 & \cellcolor[HTML]{F2F3F5}71.9 \\ \hline
 & RTN & 36.5 & 55.9 & 50.4 & 45.9 & 71.8 & 50.3 & 51.8 \\
 & GPTQ & 37.3 & 57.2 & 50.7 & 44.8 & 70.9 & 54.3 & 52.5 \\
 & AWQ & 38.7 & 58.1 & 51.0 & 44.3 & 70.8 & 55.2 & 53.0 \\
 & MBQ & 43.9 & 67.3 & 59.8 & 66.4 & 81.2 & 72.2 & 65.1 \\
 & VPTQ & 44.9 & 68.1 & 67.2 & 65.6 & 81.1 & 76.9 & 67.3 \\
 & QuIP\# & 45.6 & 69.0 & 66.9 & 66.4 & 83.1 & 78.4 & 68.2 \\
\multirow{-7}{*}{2} & \cellcolor[HTML]{F2F3F5}MGVQ & \cellcolor[HTML]{F2F3F5}46.8 & \cellcolor[HTML]{F2F3F5}68.9 & \cellcolor[HTML]{F2F3F5}74.9 & \cellcolor[HTML]{F2F3F5}67.1 & \cellcolor[HTML]{F2F3F5}83.4 & \cellcolor[HTML]{F2F3F5}79.0 & \cellcolor[HTML]{F2F3F5}70.0 \\ \hline
\end{tabular}
}
\end{table}

% Qwen2-VL-72B
\begin{table}[h]
\centering
\caption{Results of Qwen2-VL-72B. }
\label{tab:qwen2-VL-72B}
\resizebox{1.0\textwidth}{!}{
\begin{tabular}{c|c|ccccccc}
\hline
Bit & Method & \multicolumn{1}{c}{MMMU} & \multicolumn{1}{c}{SEED} & \multicolumn{1}{c}{OCRBench} & \multicolumn{1}{c}{VizWiz} & \multicolumn{1}{c}{ScienceQA} & \multicolumn{1}{c}{TextVQA} & \multicolumn{1}{c}{Average ($\uparrow$)} \\ \hline
FP16 & - & 61.1 & 77.6 & 79.9 & 76.0 & 91.6 & 82.5 & 78.1 \\ \hline
 & RTN & 57.7 & 77.5 & 70.4 & 74.8 & 89.7 & 79.7 & 75.0 \\
 & GPTQ & 57.3 & 77.2 & 78.5 & 73.6 & 91.5 & 81.6 & 76.6 \\
 & AWQ & 59.6 & 77.6 & 79.6 & 75.4 & 90.4 & 82.4 & 77.5 \\
 & MBQ & 59.6 & 77.7 & 79.4 & 75.6 & 90.5 & 82.5 & 77.6 \\
 & VPTQ & 59.4 & 77.6 & 79.0 & 75.8 & 90.9 & 82.1 & 77.5 \\
\multirow{-6}{*}{3} & \cellcolor[HTML]{F2F3F5}MGVQ & \cellcolor[HTML]{F2F3F5}60.6 & \cellcolor[HTML]{F2F3F5}77.7 & \cellcolor[HTML]{F2F3F5}79.3 & \cellcolor[HTML]{F2F3F5}75.8 & \cellcolor[HTML]{F2F3F5}91.4 & \cellcolor[HTML]{F2F3F5}82.2 & \cellcolor[HTML]{F2F3F5}77.8 \\ \hline
 & RTN & 42.1 & 66.9 & 61.3 & 62.3 & 80.2 & 71.3 & 64.0 \\
 & GPTQ & 44.2 & 68.1 & 66.3 & 65.4 & 82.5 & 75.8 & 67.1 \\
 & AWQ & 44.5 & 67.3 & 66.9 & 64.3 & 80.4 & 74.5 & 66.3 \\
 & MBQ & 48.9 & 71.4 & 74.9 & 69.8 & 82.9 & 78.4 & 71.1 \\
 & VPTQ & 53.2 & 73.4 & 76.4 & 69.1 & 83.4 & 79.8 & 72.6 \\
 & QuIP\# & 55.8 & 73.9 & 76.1 & 72.3 & 85.8 & 79.5 & 73.9 \\
\multirow{-7}{*}{2} & \cellcolor[HTML]{F2F3F5}MGVQ & \cellcolor[HTML]{F2F3F5}58.8 & \cellcolor[HTML]{F2F3F5}75.9 & \cellcolor[HTML]{F2F3F5}78.0 & \cellcolor[HTML]{F2F3F5}73.1 & \cellcolor[HTML]{F2F3F5}87.9 & \cellcolor[HTML]{F2F3F5}81.3 & \cellcolor[HTML]{F2F3F5}75.8 \\ \hline
\end{tabular}
}
\end{table}

%% file: pseudo_pseudo_SSMQ.tex
\begin{algorithm}[]
\caption{SSMQ: Sensitivity-driven Structured Mixed-precision Quantization}
\label{alg:ssmq}
\begin{algorithmic}[1]

\REQUIRE Weight matrix $W \in \mathbb{R}^{m \times n}$, total bit budget $B$
\ENSURE Quantized weights $\hat{W}$

\STATE \textbf{CSA: Channel Sensitivity Assessment}
\STATE Compute Hessian factors $H_I, H_O$ via Kronecker approximation \hfill(Eq.~\ref{H_I,H_O})
\FOR{each input/output channel}
  \STATE Compute global sensitivity (diag of $H_I, H_O$) and local sensitivity (activation norm) \hfill(Eq.~\ref{I_g}, \ref{I_l})
  \STATE Fuse normalized scores:
    $I^{(in/out)} = \log\!\bigl(\hat{I}_g^{(in/out)} \cdot \hat{I}_l^{(in/out)}\bigr)$\hfill(Eq.~\ref{I})
\ENDFOR

\STATE \textbf{MRSBP: Reordering \& Partitioning}
\STATE Sort channels by $I^{(in)}$ and $I^{(out)}$, define sensitivity $ I_{i,j} = I^{(out)}[i] \cdot I^{(in)}[j]$\hfill(Eq.~\ref{I_i_j})
\STATE Partition $W$ into four blocks and compute block sensitivity $S_t$\hfill(Eq.~\ref{sensitivity_bit_efficiency_ratio})

\STATE \textbf{OBA: Optimal Bit Allocation}
\FOR{each block $t$}
  \STATE $b_t = B \cdot \frac{\sqrt{S_t}}{\sum_{s=1}^{4} \sqrt{S_s}}$
  , then quantize block with $b_t$ bits
  \STATE \hfill(Eq.~\ref{b_t})
\ENDFOR

\RETURN $\hat{W}$

\end{algorithmic}
\end{algorithm}

%% file: pseudo_pseudo_GAEC.tex
\begin{algorithm}[]
\caption{GAEC: Gradient-Aware Error Compensation}
\label{alg:gaec}
\begin{algorithmic}[1]

\REQUIRE Original weights $W$, Hessian blocks $H_O, H_I$, quantizer $\mathcal{Q}$,
scaling factor $\beta$, tolerance $\varepsilon$, maximum iterations $K$
\ENSURE Optimized quantized weights $\hat{W}$

\STATE \textbf{Initialization}
\STATE $(L_O, D_O) = \text{BlockLDL}(H_O)$
\STATE $(L_I, D_I) = \text{BlockLDL}(H_I)$
\STATE $\hat{W} = \mathcal{Q}(W)$; $t = 0$

\STATE \textbf{Iterative Compensation}
\WHILE{$t < K$ and not converged}
  \STATE $E^{(t)} = W - \hat{W}^{(t)}$%(see Eq.~\ref{E})
  \STATE 
    $A^{(t)} = (L_O + I)^{-T} \, E^{(t)} \, (L_I + I)^{-1}$ \hfill({Eq.~\ref{A}})
  \STATE 
    $T = \beta A^{(t)}.$\hfill({Eq.~\ref{T}})
  \STATE
    $\eta = W + L_O^T E^{(t)} L_I + L_O^T E^{(t)} + E^{(t)} L_I - T$ \hfill({Eq.~\ref{eta}})
  \STATE $\hat{W}_{\text{new}} = \mathcal{Q}(\eta)$\hfill({Eq.~\ref{w_hat_new}})
  \STATE 
   $\Delta = \frac{\lVert \hat{W}_{\text{new}} - \hat{W} \rVert_F}
    {\max(1, \lVert \hat{W} \rVert_F)}$ 
  
  \IF{$\Delta < \varepsilon$}
    \STATE break
  \ENDIF
  \STATE $\hat{W} = \hat{W}_{\text{new}}$; $t = t + 1$
\ENDWHILE
\RETURN $\hat{W}$
\end{algorithmic}
\end{algorithm}